\documentclass[10pt,twocolumn,letterpaper]{article}

\usepackage{cvpr}              

\usepackage{graphicx}
\usepackage{amsmath}
\usepackage{amssymb}
\usepackage{booktabs}

\usepackage{enumitem}
\usepackage{verbatim}
\usepackage{pifont}
\usepackage[pagebackref,breaklinks,colorlinks]{hyperref}
\usepackage{amssymb}
\usepackage{dblfloatfix}
\usepackage{footmisc}
\usepackage{graphicx}
\usepackage{booktabs}
\usepackage{float}
\usepackage{dblfloatfix}

\usepackage{multirow}

\usepackage{pifont}
\newcommand{\xmark}{\ding{55}}%
\newcommand{\cmark}{\ding{51}}%
\usepackage[pagebackref,breaklinks,colorlinks]{hyperref}

\usepackage[capitalize]{cleveref}
\crefname{section}{Sec.}{Secs.}
\Crefname{section}{Section}{Sections}
\Crefname{table}{Table}{Tables}
\crefname{table}{Tab.}{Tabs.}

\def\cvprPaperID{6014} 
\def\confName{CVPR}
\def\confYear{2023}

\begin{document}

\title{  Instance Relation Graph Guided \\ Source-Free Domain Adaptive Object Detection}

\author{Vibashan VS, Poojan Oza, and Vishal M. Patel \\
 Johns Hopkins University, Baltimore, MD, USA \\
{\tt\small \{vvishnu2,poza2,vpatel36\}@jhu.edu}
}

\maketitle

\begin{abstract}
Unsupervised Domain Adaptation (UDA) is an effective approach to tackle the issue of domain shift. Specifically, UDA methods try to align the source and target representations to improve generalization on the target domain. Further, UDA methods work under the assumption that the source data is accessible during the adaptation process. However, in real-world scenarios, the labelled source data is often restricted due to privacy regulations, data transmission constraints, or proprietary data concerns. The Source-Free Domain Adaptation (SFDA) setting aims to alleviate these concerns by adapting a source-trained model for the target domain without requiring access to the source data. In this paper, we explore the SFDA setting for the task of adaptive object detection. To this end, we propose a novel training strategy for adapting a source-trained object detector to the target domain without source data. More precisely, we design a novel contrastive loss to enhance the target representations by exploiting the objects relations for a given target domain input. These object instance relations are modelled using an Instance Relation Graph (IRG) network, which are then used to guide the contrastive representation learning. In addition, we utilize a student-teacher to effectively distill knowledge from source-trained model to target domain. Extensive experiments on multiple object detection benchmark datasets show that the proposed approach is able to efficiently adapt source-trained object detectors to the target domain, outperforming state-of-the-art domain adaptive detection methods. Code and models are provided in \href{https://viudomain.github.io/irg-sfda-web/}{https://viudomain.github.io/irg-sfda-web/}.

\vspace{-3.0mm}
\end{abstract}

\vspace{-3.0mm}

\section{Introduction}
\vskip -10.0pt

\label{sec:intro}

\begin{figure}[t!]
    \begin{center}
        \includegraphics[width=1.0\linewidth]{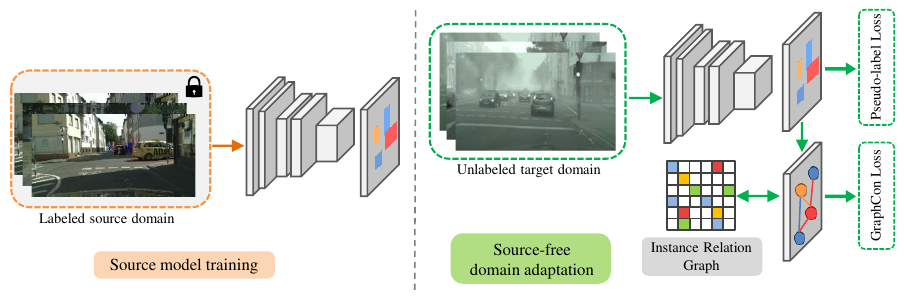}
    \end{center}
\vskip -20.0pt \caption{\textbf{Left:} Supervised training of detection model on the source domain. \textbf{Right:} Source-Free Domain Adaptation (SFDA) setup, i.e., the source-trained model is adapted to the target domain in the absence of source data with pseudo-label self-training and proposed Instance Relation Graph (IRG) network guided contrastive loss.}
    \label{fig:intro} 
\vskip -20.0pt
\end{figure}

In recent years, object detection has seen tremendous advancements due  to the rise of deep networks \cite{zou2019object,liu2020deep,redmon2016you,liu2016ssd,lin2017focal,duan2019centernet}.
The major contributor to this success is the availability of large-scale annotated  detection datasets \cite{everingham2010pascal,cordts2016cityscapes,geiger2013vision,lin2014microsoft,yu2020bdd100k}, as it enables the supervised training of deep object detector models.
However, these models often have poor generalization when deployed in visual domains not encountered during training.
In such cases, most works in the literature follow the Unsupervised Domain Adaptation (UDA) setting to improve generalization \cite{ganin2016domain,tzeng2017adversarial,saito2018maximum,lo2022learning,hoffman2018cycada,chen2017no,hoffman2016fcns}.
Specifically, UDA methods aim to minimize the domain discrepancy by aligning the feature distribution of the detector model between source and target domain \cite{chen2018domain,saito2019strong,inoue2018cross,he2019multi,Sindagi_DA_Detection_ECCV2020}.
To perform feature alignment, UDA methods require simultaneous access to the labeled source and unlabeled target data.
However in practical scenarios, the access to source data is often restricted due to concerns related to privacy/safety, data transmission, data proprietary etc.
For example, consider a detection model trained on large-scale source data, that performs poorly when deployed in new devices having data with different visual domains.
In such cases, it is far more efficient to transmit the source-trained detector model ($\sim$500-1000MB) for adaptation rather than transmitting the source data ($\sim$10-100GB) to these new devices \cite{kundu2020universal,huang2021model}.
Moreover, transmitting only source-trained model alleviates many privacy/safety, data proprietary concerns as well \cite{liu2021source,xia2021adaptive,liang2020we}.
Hence, \textit{adapting the source-trained model to the target domain without having access to source data is essential} in the case of practical deployment of detection models.
This motivates us to study Source-Free Domain Adaptation (SFDA) setting for adapting object detectors (illustrated in Fig.~\ref{fig:intro}).

The SFDA is a more challenging setting than UDA.
Specifically, on top of having no labels for the target data, the source data is not accessible during adaptation.
Therefore, most SFDA methods for detection consider training with pseudo-labels generated by source-trained model \cite{li2020free,huang2021model}.
However, these pseudo-labels are noisy due to domain shift and the training on noisy pseudo-label is sub-optimal solution \cite{liu2021unbiased,deng2021unbiased}. 
In order to avoid such a scenario, we need to consider two challenges of the SFDA training, \textit{1) Effectively distill target domain information into source-trained model and 2) Enhancing the target domain feature representations}.

\begin{figure}[t]
\centering
\begin{center}
        \includegraphics[width=0.48\linewidth]{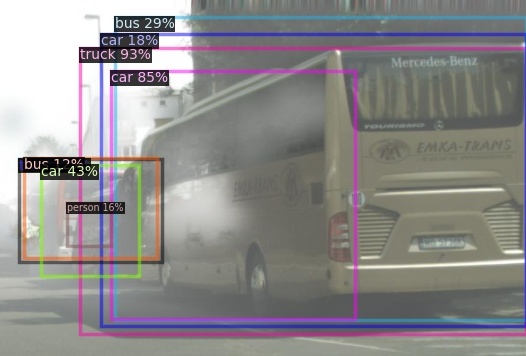}
        \includegraphics[width=0.48\linewidth]{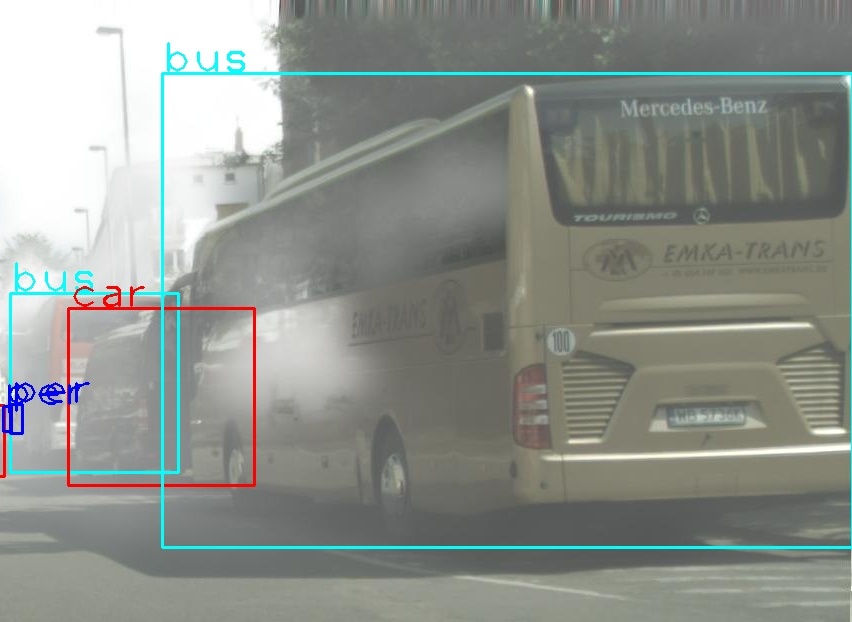}\\
        \hskip 10.5pt \small{(a) Prediction} \hskip 35pt \small{(b) Ground truth} 
\end{center}
\vskip -18.0pt \caption{\small (a) Object predictions by Cityscapes-trained model on the FoggyCityscapes  image. (b) Corresponding ground truth. Here, the proposals around the bus instance have inconsistent predictions, indicating that instance features are prone to large shift in the feature space, for a small shift in the proposal location.}
\label{fig:gt_comp}
\vskip -18.0pt
\end{figure}

A critical challenge is improving the features of the target domain data.
Consider Fig.~\ref{fig:gt_comp}, which shows object proposals for an image from FoggyCityscapes \cite{Sakaridis2018SemanticFS}, predicted by a detector model trained on Cityscapes \cite{cordts2016cityscapes}.
Here, all the proposals have Intersection-over-Union$>$0.9 with respective ground-truth bounding boxes and each proposal is assigned a prediction with a confidence score.
Noticeably, the proposals around the bus instance have different predictions, e.g., car with 18$\%$, truck with 93$\%$, and bus with 29$\%$ confidence.
This indicates that the pooled features are prone to a large shift in the feature space for a small shift in the proposal location.
This is because, the source-trained model representations would tend to be biassed towards source data, resulting in weak representation for the target data.
To this end, we utilize the Contrastive Representation Learning (CRL) framework \cite{chen2020simple,he2020momentum,chopra2005learning,khosla2020supervised} and design a novel contrastive loss to enhance the feature representations of the target domain.

\begin{figure}[t!]
    \begin{center}
        \includegraphics[width=0.95\linewidth]{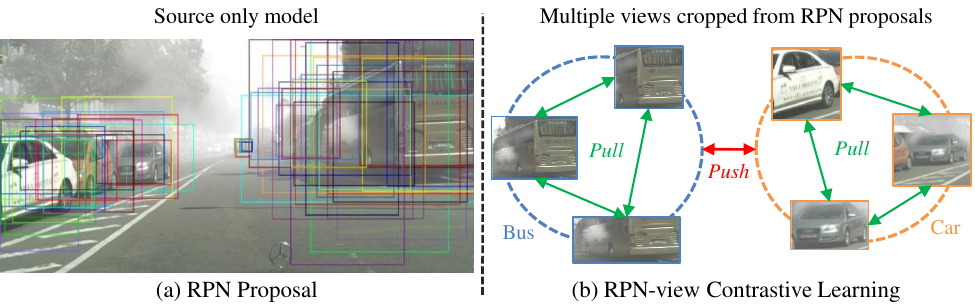}
    \end{center}
 \vskip -20.0pt \caption{(a) Class agnostic object proposals generated by Region Proposal Network (RPN). (b) Cropping out RPN proposals will provide multiple contrastive views of an object instance.  We  utilize this to improve target domain feature representations through RPN-view contrastive learning. However as RPN proposals are class agnostic, it is challenging to form positive (same class)/negative pairs (different class), which is essential for CRL.}
    \label{fig:motive_contrastive_learning} 
\vskip -17.0pt
\end{figure}

CRL has been shown to learn high-quality representations from images in an unsupervised manner \cite{chen2020simple,chen2020big,wu2018unsupervised}.
CRL methods achieve this by forcing representations to be similar under multiple views (or augmentations) of an anchor image and dissimilar to all other images. 
In classification, the CRL methods assume that each image contains only one object.
On the contrary, for object detection, each image is highly likely to have multiple object instances.
Furthermore, the CRL training also requires large batch sizes and multiple views to learn high-quality representations, which incurs a very high GPU/memory cost, as detection models are computationally expensive.
To circumvent these issues, we propose an alternative strategy which exploits the architecture of the detection model like Faster-RCNN \cite{ren2015faster}.
Interestingly, the proposals generated by the Region Proposal Network (RPN) of a Faster-RCNN essentially provide multiple views for any object instance as shown in Fig.~\ref{fig:motive_contrastive_learning} (a).
In other words, \textit{the RPN module provides instance augmentation for free}, which could be exploited for CRL, as shown in Fig.~\ref{fig:motive_contrastive_learning} (b).
However, RPN predictions are class agnostic and without the ground-truth annotations for target domain, it is impossible to know which of these proposals would form positive (same class)/negative pairs (different class), which is essential for CRL.
To this end, we propose a Graph Convolution Network (GCN) based network that models the inter-instance relations for generated RPN proposals.
Specifically, each node corresponds to a proposal and the edges represent the similarity relations between the proposals.
This learned similarity relations are utilized to extract information regarding which proposals would form positive/negative pairs and are used to guide CRL.
By doing so, we show that such graph-guided contrastive representation learning is able to enhance representations for the target data.

Our contributions are summarized as follows:
\begin{itemize}[topsep=0pt,noitemsep,leftmargin=*]
 	\item We investigate the problem of source-free domain adaptation for object detection and identify some of the major challenges that need to be addressed.
 	\item We introduced an Instance Relation Graph (IRG) framework to model the relationship between proposals generated by the region proposal network.
 	\item We propose a novel contrastive loss which is guided by the IRG network to improve the representations for the target data.
 	\item The effectiveness of the proposed method is evaluated on multiple object detection benchmarks comprising of visually distinct domains. Our method outperforms existing source-free domain adaptation methods and many unsupervised domain adaptation methods.
 \end{itemize}



\section{Related works}\label{sec:related_works}
\vskip -8.0pt

\noindent \textbf{Unsupervised Domain Adaption.} Unsupervised domain adaptation for object detection was first explored by Chen \etal \cite{Chen2018DomainAF}. Chen \etal \cite{Chen2018DomainAF} proposed adversarial-based feature alignment for a Faster-RCNN network at image and instance level to mitigate the domain shift. Later, Saito \etal \cite{saito2019strong} proposed a method that performs strong local feature alignment and weak global feature alignment based on adversarial training. Instead of utilizing an adversarial-based approach,  Khodabandeh \etal \cite{khodabandeh2019robust} proposed to mitigate domain shift by pseudo-label self-training on the target data. Self-training using pseudo-labels ensures that the detection model learns target representation. Later, Kim \etal \cite{kim2019diversify} proposed an image-to-image generation based adaptation strategy where given source and target domain, the proposed method generates target like source images. The generated target-like images are then used to train the detection model; as a result, the detection network learn target features. Recently, Hsu \etal \cite{hsu2020every} explored domain adaptation for one-stage object detection, where he utilized a one-stage object detection framework to perform object center-aware training while performing adversarial feature alignment. There exists multiple UDA work for object detection \cite{cai2019exploring,he2019multi,Sindagi_DA_Detection_ECCV2020, roychowdhury2019automatic,nair2021confidence,wu2021instance,vs2022meta}; however, all these works assume you have access to labeled source and unlabeled target data.

\noindent \textbf{Source-Free Domain Adaptation.} In a real-world scenario, the source data is not often accessible during the adaptation process due to privacy regulations, data transmission constraints, or proprietary data concerns. Many works have addressed the source-free domain adaptation (SFDA) setting for classification \cite{liang2020we,li2020model}, 2D and 3D object detection \cite{vs2023towards, hegde2021attentive,hegde2021uncertainty,huang2021model} and video segmentation \cite{lo2023spatio} tasks.   First for the classification task, the SFDA setting was explored by Liang \etal \cite{liang2020we} proposed source hypothesis transfer, where the source-trained model classifier is kept frozen and target generated features are aligned via pseudo-label training and information maximization. Following the segmentation task Liu \etal \cite{liu2021source} proposed a self-supervision and knowledge transfer-based adaptation strategy for target domain adaptation.  For object detection task, \cite{li2020free} proposed a pseudo-label self-training strategy and \cite{huang2021model} proposed self-supervised feature representation learning via previous models approach.

\noindent \textbf{Contrastive Learning.} The huge success in unsupervised feature learning is due to contrastive learning which has attributed to huge improvement in many unsupervised tasks \cite{oord2018representation, chen2020simple, huang2021model}.  Contrastive learning generally learns a discriminative feature embedding by maximizing the agreement between positive pairs and minimizing the agreement with negative pairs. In \cite{oord2018representation,chen2020simple,he2020momentum}. in batch of an image, an anchor image undergoes different augmentation and these augmentations for that anchor forms positive pair and negative pairs are sampled from other images in the given batch. Later, in \cite{khosla2020supervised} exploiting the task-specific semantic information, intra-class features embedding is pulled together and repelled away from cross-class feature embedding. In this way, \cite{khosla2020supervised} learned a more class discriminative feature representation. All these works are performed for the classification task, and these methods work well for large batch size tasks \cite{chen2020simple, khosla2020supervised}. Extending this to object detection tasks generally fails as detection models are computationally expensive. To overcome this, we exploit graph convolution networks to guide contrastive learning for object detection.

\noindent{\bf{Graph Convolution Neural Networks (GNNs).}} Graph Convolution Neural Networks was first introduced by  Gori  \cite{gori2005new} to process the data with a graph structure using neural networks. The key idea is to construct a graph with nodes and edges relating to each other and update node/edge features, i.e., a process called node feature aggregation. In recent years, different GNNs have been proposed (e.g., GraphConv \cite{morris2019weisfeiler}, GCN \cite{kipf2016semi}, each with a unique feature aggregation rule which is shown to be effective on various tasks. Recent works in image captioning \cite{zhong2020comprehensive,nguyen2021defense}, scene graph parsing \cite{yang2018graph} etc. try to model inter-instance relations by IoU based graph generation.  For these applications, IoU based graph is effective as modelling the interaction between objects is essential and can be achieved by simply constructing a graph based on object overlap. However, the problem araises with IoU based graph generation when two objects have no overlap and in these cases, it disregards the object relation. For example, see Fig. \ref{fig:motive_contrastive_learning} (a), where the proposals for the left sidecar and right sidecar has no overlap; as a result, IoU based graph will output no relation between them. In contrast for the CRL case, they need to be treated as a positive pair. To overcome these issues, we propose a learnable graph convolution network to models inter-instance relations present within an image.  


\section{Proposed method}\label{sec:proposed_method}
\vskip -8.0pt
\subsection{Preliminaries}
\noindent \textbf{Background.} UDA \cite{tzeng2017adversarial,hoffman2016fcns,chen2018domain} considers labeled source and unlabeled target domain datasets for adaptation.
Let us formally denote the labeled source domain dataset as ${D}_s = \{x_s^n, y_s^n\}_{n=1}^{N_s}$, where $x_s^n$ denotes the $n^{th}$ source image and $y_s^n$ denotes the corresponding ground-truth, and the unlabeled target domain dataset as, ${D}_t = \{x_t^n\}_{n=1}^{N_t}$, where $x_t^n$ denotes $n^{th}$ the target image without the  ground-truth annotations.
In contrast, the SFDA setting \cite{liang2020we,li2020free,kim2021domain,liu2021source} considers a more practical scenario where the access to the source dataset is restricted and only a source-trained model $\Theta$ and the unlabeled target data ${D}_t$ are available during adaptation.

 \begin{figure*}[t!]
 	\begin{center}
        \includegraphics[width=0.85\linewidth]{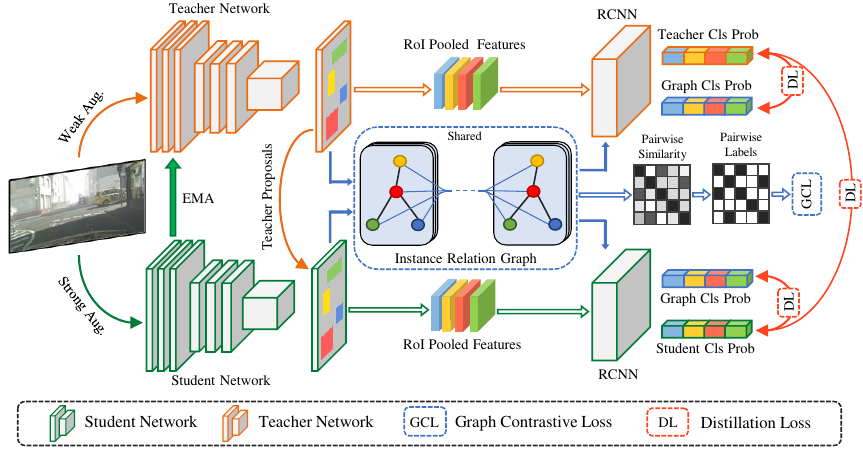}
 	\end{center}
 \vskip -15.0pt \caption{Overall architecture of our method. We follow a student-teacher framework for the detector model training. The proposed Instance Relation Graph (IRG) network models the relation between the object proposals generated by the detector. Using the inter-proposal relations learned by IRG, we generate pairwise labels to identify positive/negative pairs for contrastive learning. The IRG network is regularized with distillation loss between student and teacher model.}
 	\label{fig:proposed_method_block_diagram} 
\vskip -15.0pt
 \end{figure*}
 

\noindent \textbf{Mean-teacher based self-training.} Self-training adaptation strategy updates the model on unlabeled target data using pseudo labels generated by the source-trained model.
The pseudo labels are filtered through confidence threshold and the reliable ones are used to supervise the detector training \cite{khodabandeh2019robust}.
More formally, the pseudo label supervision loss for the object detection model can be given as:
\setlength{\belowdisplayskip}{1pt} \setlength{\belowdisplayshortskip}{1pt}
\setlength{\abovedisplayskip}{1pt} \setlength{\abovedisplayshortskip}{1pt}
\begin{multline}
\label{eq:pl_st}
\mathcal{L}_{SL} = \mathcal{L}_{cls}^{rpn}(x^{n}_{t},\tilde{y}^{n}_{t})+\mathcal{L}_{reg}^{rpn}(x^{n}_{t},\tilde{y}^{n}_{t})\\
 +\mathcal{L}_{cls}^{roi}(x^{n}_{t},\tilde{y}^{n}_{t})+\mathcal{L}_{reg}^{roi}(x^{n}_{t},\tilde{y}^{n}_{t}), 
\end{multline}
where $\tilde{y}^{n}_{t}$ is the pseudo label obtained after filtering low confident predictions.
Even after filtering out low confidence predictions, the pseudo labels generated by the source-trained model are still noisy due to the domain shift. Therefore, to effectively distill knowledge from a source-trained model, the pseudo-labels quality need to be improved \cite{liu2021unbiased,deng2021unbiased}.

To this end, we utilize mean-teacher \cite{tarvainen2017mean} which consists of student and teacher networks with parameters  $\Theta_{s}$ and $\Theta_{t}$, respectively.
In the mean-teacher, the student is trained with pseudo labels generated by the teacher and the teacher is progressively updated via Exponential Moving Average (EMA) of student weights. 
Furthermore, motivated by semi-supervised techniques \cite{liu2021unbiased,deng2021unbiased}, the student and teacher networks are fed with strong and weak augmentations, respectively and consistency between their predictions improves detection on target data.
Hence, the overall student-teacher self-training based object detection framework updates can be formulated as:
\begin{align}
\label{eq:teach_up}
\Theta_{s} &\leftarrow \Theta_{s} + \gamma \frac{\partial(\mathcal{L}_{SL}^{st})}{\partial \Theta_{s}}, \\
\Theta_{t} &\leftarrow \alpha \Theta_{t}+(1-\alpha)\Theta_{s},
\end{align}
where $\mathcal{L}_{SL}^{st}$ is the student loss computed using the pseudo-labels generated by the teacher network.
The hyperparameters $\gamma$ and $\alpha$ are student learning rate and teacher EMA rate, respectively.
Although the student-teacher framework enables knowledge distillation with noisy pseudo-labels, it is not sufficient to learn high-quality target features, as discussed earlier.
Hence, to enhance the features in the target domain, we utilize contrastive representation learning. 

\noindent \textbf{Contrastive Representation Learning (CRL).} SimCLR \cite{chen2020simple} is a commonly used CRL framework, which learns representations for an image by maximizing agreement between differently augmented views of the same sample via a contrastive loss.
More formally, given an anchor image $x_i$, the SimCLR loss can be written as:
\begin{equation}
\label{eq:simclr}
\mathcal{L}_{\text{SimCLR}} = - {\log}\left( \frac{\operatorname{exp}(\operatorname{sim}(r_i, r_j))}{\sum_{k =1, \ni_{k\neq i}}^{2N}  \operatorname{exp}(\operatorname{sim}(r_i, r_k))} \right),
\end{equation}
where $N$ is the batch size, $r_i$ and $r_j$ are the features of two different augmentations of the same sample $x_i$, whereas $r_k$ represents the feature of $k^{th}$ batch sample $x_k$, where $k \neq i$.
Also, $\operatorname{sim}(\cdot, \cdot)$ indicates a similarity function, e.g. cosine similarity. 
Note that, in general the CRL framework assumes that each image contains one category \cite{chen2020simple}.
Moreover, it requires large batch sizes that could provide multiple positive/negative pairs for the training \cite{chen2020big}.

\subsection{Graph-guided contrastive learning}

To overcome the challenges discussed earlier, we exploit the architecture of Faster-RCNN to design a novel contrastive learning strategy as shown in Fig. \ref{fig:proposed_method_block_diagram}.
As we discussed in Sec.~\ref{sec:intro}, RPN by default, provides augmentation for each instance in an image.
As shown in Fig.~\ref{fig:motive_contrastive_learning}, cropping out the RPN proposals will provide multiple different views around each instance in an image.
This property can be exploited to learn contrastive representation by maximizing the agreement between proposal features for the same instance and disagreement of the proposal features for different instances.
However, RPN predictions are class agnostic and the unavailability of ground truth boxes for target domain makes it difficult to know which proposals belong to which instance.
Consequently, for a given proposal as an anchor, sampling positive/ negative pairs become a challenging task.
To this end, we introduce an Instance Relation Graph (IRG) network that models inter-instance relations between the RPN proposals.
IRG then provides pairwise labels by inspecting similarities between two proposals to identify positive/negative proposal pairs.

\begin{figure}[b!]
    \begin{center}
 	    \includegraphics[width=1.0\linewidth]{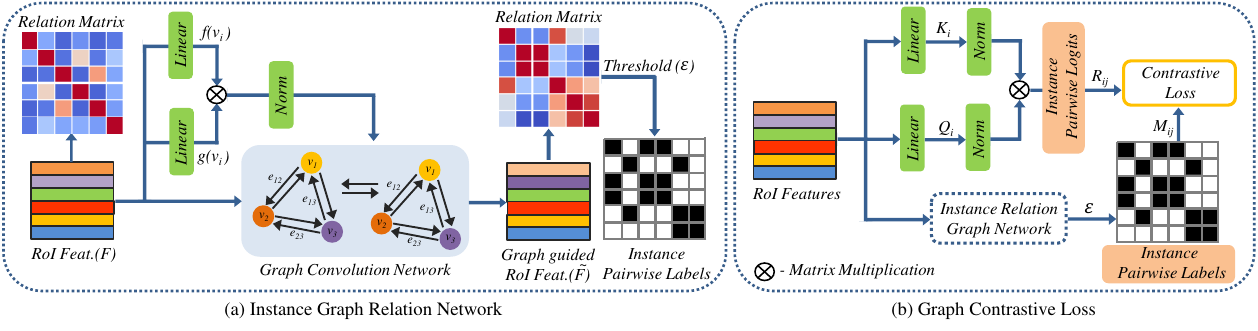}
    \end{center}
 \vskip -15pt \caption{ \textbf{(a) Instance Graph Relation Network:} Given proposal RoI features, the IRG models and improves the similarity relations between proposals. Thresholding the learned relation matrix generates instance pairwise labels used to obtain positive (white)/negative (black) pairs for computing the contrastive loss. \textbf{(b) Graph Contrastive Loss:} Projecting RoI features as keys and queries and performing transpose multiplication provides instance pair wise logits. The generated instance pairwise logits and instance pairwise labels are used to compute the contrastive loss. }
    \label{fig:IRG} 
\vskip -10pt
\end{figure}

\subsubsection{Instance Relation Graph (IRG)}

Graph Convolution Network (GCN) is an effective way to understand the relationship and propagate information between the nodes \cite{cai2018comprehensive,wu2019simplifying,zhang2020deep}.
The proposed IRG network utilizes GCN to learn the relationship between the RPN proposals.
Let us denote IRG as $\mathcal{G} : \mathcal{G} =\langle \mathcal{V} , \mathcal{E} \rangle$, where $\mathcal{V}$  is nodes and $\mathcal{E}$  is edges of the graph network. The nodes in $\mathcal{V}$ corresponds to RoI features extracted from RPN proposals and edges $e_{i,j} \in \mathcal{E}$ encodes relationship between the $i^{th}$ and the $j^{th}$ proposals.
We then aim to learn relation matrix $\mathcal{E}$, to find the relationship between the RPN proposals.
Both the student and teacher networks share the IRG network for modeling relationships between object proposals.

\noindent \textbf{Nodes.} The nodes in IRG represent features of the RPN proposals obtained from RoI feature extractor.
The nodes in $\mathcal{G}$ are denoted as $\mathcal{V}=\{v_{1}, v_{2}, ..., v_{m}\}$, where $v_{m}$ is the feature of the $m^{th}$ instance.
Here, $m$ is the total number of RPN proposals.
We set $m$ to 300 for both teacher and student.
The teacher pipeline has input with weak augmentations; thus, the teacher RPN proposals are better and more consistent than strongly augmented student RPN.
Hence, we use teacher RPN proposals to extract RoI features and construct IRG for both student and teacher networks.

\noindent \textbf{Edges.} The edges in the graph $\mathcal{G}$ are denoted as $\mathcal{E} = [e_{ij}]_{m \times m}$, where $e_{ij}$ is the edge of the $v_i^{th}$ and $v_j^{th}$ nodes, denoting the relation of corresponding instances in the feature space and can be formally represented as: 
\begin{equation}\label{eq:edge}
	e_{ij} = \frac{\operatorname{exp}(S_{ij})}{\sum_{}^{}\operatorname{exp}(S_{ij})}, \text{ where } S_{ij}=f(v_i) \cdot g(v_j)^{T},
\end{equation}
where, $f$ and $g$ are learnable function helps to model relation between nodes.

\noindent \textbf{Graph Distillation Loss (GDL).} Let us denote the input features to IRG as $F \in \mathbb{R}^{m \times d}$ where $m$ denoting the number of proposal instances and  $d$ denoting the feature dimension of the RoI features.
The features $F$ are then passed through graph convolution layers of IRG to model the inter-instance relations.
The output features $\tilde{F}$ are calculated as:
\begin{equation}\label{eq:edge}
    \tilde{F} = \operatorname{ReLU}(\mathcal{E} F {W}),
\end{equation}
where $W$ is a learnable weight matrix.
Subsequently, both features ${F}$ and $\tilde{F}$ are fed into the RCNN classification layer to obtain class logits for each proposal.
Let us denote the student and the teacher class logits corresponding to features ${F}$ as $Z_{st}$ and $Z_{te}$, respectively.
Similarly, let us denote student and teacher class logits corresponding to IRG output features $\tilde{F}$ as $\tilde{Z}_{st}$ and $\tilde{Z}_{te}$, respectively.

To supervise the IRG network parameters, we minimize the discrepancy between class logits $Z$ and $\tilde{Z}$ for both student and teacher pipeline in an end-to-end manner.
In addition, we also minimize the discrepancy between student and teacher class logits ${Z}_{st}$ and ${Z}_{te}$ to maintain consistency between both pipelines.
We denote this discrepancy as GDL which can be formally written as:
\begin{align*}\label{eq:overall_objective}
\mathcal{L}_{GDL}  &=  \operatorname{KL}(\sigma(Z_{st}),\sigma(\tilde{Z}_{st})) \\&+   \operatorname{KL}(\sigma(Z_{te}),\sigma(\tilde{Z}_{te}))  +   \operatorname{KL}(\sigma(Z_{st}),\sigma(Z_{te})),
\end{align*}
where $\operatorname{KL}$ denotes the Kullback–Leibler divergence, $\sigma$ denotes softmax operator. Therefore, minimizing $\mathcal{L}_{GDL}$ supervises the IRG network which inturn learns the instance relation matrix ($\mathcal{E}$).

\vspace{-1. em}

\subsubsection{Graph Contrastive Loss (GCL)} 
\vspace{-1.5 em}
\noindent \\\textbf{Instance pairwise labels.} In order to utilize the contrastive loss, we need to understand the relation of the given anchor proposal with other RPN proposals to form positive/negative pairs. As mentioned earlier, this relation matrix ($\mathcal{E}$) is obtained from the IRG network, which learns how proposals are related to each other. For instance pairwise label generation, let us consider proposal instances $i$ and $j$ and it's corresponding learned relation between them, $e_{ij} \in \mathcal{E}$. Now, one can obtain positive/negative pairs by simply setting a threshold $\epsilon$ on normalized $\mathcal{E}$ where the $e_{ij} > \epsilon$ would indicate that they are highly related, forming a positive pair and vice versa for the negative pairs.
The pairwise labels between $i^{th}$ and $j^{th}$ proposal instances, denoted as $M_{ij}$, can be given as:
\begin{equation}
  M_{ij} =
    \begin{cases}
      0, \ \ e_{ij} < \epsilon \\
      1, \ \ e_{ij}  \ \geq \epsilon,
    \end{cases}       
\end{equation}
where $\epsilon$ is a hyper parameter. Thus, for a given anchor proposal we obtain its corresponding positive and negative proposal pairs from $M_{ij}$ . 

\noindent \textbf{Instance pairwise logits.} As shown in Fig. \ref{fig:IRG}, the RoI features $v_{i}$ are projected as key $k_{i}$  and query $q_{i}$ inorder to model better correlation among the RoI features\cite{vaswani2017attention}. For given $i^{th}$ RoI features, we obtain key, query and pairwise logits as follows: 
\begin{align*}
\label{eqn:eqlabel}
\begin{split}
 k_i &= W_k \cdot v_i,
\\
 q_i &= W_q \cdot v_i,
 \\
 R_{ij} &= q_i(k_j)^T,
\end{split}
\end{align*}
where $W_k$ and $W_q$ are linear layer weights and $k_{i}$, $q_{i}$ and $R_{ij}$ are key, query and instance pairwise logits. To this end, the contrastive loss can be computed from the instance pairwise logits ($R_{ij}$) and instance pairwise labels ($M_{ij}$).

\noindent \textbf{Contrastive loss.} Considering any $i^{th}$ proposal as an anchor, where $i \in I \equiv \left\{ 1, 2,...,m \right\}$, let us define a set consisting of all the samples excluding the anchor as $A(i) \equiv I \backslash  \left\{ i \right\}$.
Further, using pairwise labels from $M$, we can create a positive pair set defined as $P(i) \equiv  \left\{ p\in I : M_{ij} = 1\right\} \backslash  \left\{ i \right\}$.
For given $i^{th}$ proposal, the Graph Contrastive Loss (GCL) can be calculated as: 
\begin{small}
\begin{equation}
\mathcal{L}_{GCL} = \sum_{i\in I}^{} -\log\left\{ \frac{1}{\left| P(i) \right|} \sum_{p\in P(i)}^{} \frac{\operatorname{exp} \left( q_i(k_p)^{T} \right)}{ \sum_{a\in A(i)}^{}\operatorname{exp} \left( q_i(k_a)^{T} \right) }   \right\},     
\end{equation}
\end{small}
By training with the proposed loss $\mathcal{L}_{GCL}$, the student network is encouraged to learn high-quality feature representations on the target domain.  We show that it improves the detector's performance by conducting experimental analysis in Sec.~\ref{sec:experiments}.  Note that GCL is used only to update the student network parameters, whereas the teacher network parameters are updated via EMA.

\vspace{-0.5 em}

\begin{table}[!t]
\caption{Quantitative results (mAP) for Cityscapes $\rightarrow$  FoggyCityscapes. S: Source only, O: Oracle, UDA: Unsupervised Domain Adaptation, SFDA: Source-Free Domain Adaptation.}
\label{tab:foggy}
\centering
\resizebox{1.00\linewidth}{!}{\begin{tabular}{clccccccccc} \hline
Type  & Method         & prsn          & rider         & car           & truck         & bus           & train         & mcycle        & bicycle       & mAP           \\ \hline
\hline
S  & Source Only    & 29.3          & 34.1          & 35.8          & 15.4          & 26.0          & 9.09          & 22.4          & 29.7          & 25.2          \\ \hline
 & DA Faster \cite{Chen2018DomainAF}     & 25.0          & 31.0          & 40.5          & 22.1 & 35.3          & 20.2          & 20.0          & 27.1          & 27.6          \\ 
 & D\&Match \cite{kim2019diversify}  & 30.8& 40.5& 44.3& 27.2& 38.4& 34.5& 28.4& 32.2& 34.6\\
 & MAF \cite{he2019multi} (ICCV 2019)& 28.2& 39.5& 43.9& 23.8& 39.9& 33.3& 29.2& 33.9& 34.0\\
 & Robust DA \cite{khodabandeh2019robust} (ICCV 2019) & 35.1& 42.1& 49.1& 30.0& 45.2& 26.9& 26.8& 36.0& 36.4\\
 & MTOR \cite{cai2019exploring}  & 30.6& 41.4& 44.0& 21.9& 38.6& 40.6& 28.3& 35.6& 35.1\\
UDA & SWDA \cite{saito2019strong}  & 29.9& 42.3& 43.5& 24.5& 36.2& 32.6& 30.0& 35.3& 34.3\\
  & CDN \cite{su2020adapting} & 35.8& 45.7& 50.9& 30.1& 42.5& 29.8& 30.8& 36.5& 36.6\\
  & Collaborative DA \cite{zhao2020collaborative}  & 32.7& 44.4& 50.1& 21.7& 45.6& 25.4& 30.1& 36.8& 35.9\\
  & iFAN DA \cite{zhuang2020ifan}  & 32.6& 48.5& 22.8& 40.0& 33.0& 45.5& 31.7& 27.9& 35.3\\
  & Instance DA \cite{zhuang2020ifan} & 33.1& 43.4& 49.6& 21.9& 45.7& 32.0& 29.5& 37.0& 36.5\\
    & Progressive DA \cite{hsu2020progressive} & 36.0& 45.5& 54.4& 24.3& 44.1& 25.8& 29.1& 35.9& 36.9\\ 
    & Categorical DA \cite{xu2020exploring} & 32.9& 43.8& 49.2& 27.2& 45.1& 36.4& 30.3& 34.6& 37.4\\
    & MeGA CDA \cite{hsu2020progressive} & 37.7& 49.0& 52.4& 25.4& 49.2& 46.9& 34.5& 39.0& 41.8\\
    & Unbiased DA \cite{deng2021unbiased}  & 33.8& 47.3& 49.8& 30.0& 48.2& 42.1& 33.0& 37.3& 40.4\\\hline
  & SFOD \cite{li2020free} & 21.7& 44.0& 40.4& 32.2& 11.8& 25.3& 34.5& 34.3& 30.6\\
 & SFOD-Mosaic \cite{li2020free} & 25.5& 44.5& 40.7& \textbf{33.2}& 22.2& \textbf{28.4}& 34.1& 39.0& 33.5\\
 SFDA & HCL \cite{huang2021model} & 26.9& \textbf{46.0}& 41.3& 33.0& 25.0& 28.1& \textbf{35.9}& 40.7& 34.6\\
 & LODS \cite{li2022source} & 34.0& 45.7& 48.8& 27.3& 39.7& 19.6& 33.2& 37.8& 35.8\\
  & Mean-Teacher \cite{tarvainen2017mean} & 33.9  & 43.0& 45.0& 29.2& 37.2& 25.1& 25.6& 38.2 & 34.3  \\ 
  & IRG (Ours)    & \textbf{37.4}& 45.2& \textbf{51.9}& 24.4& \textbf{39.6}& 25.2& 31.5&  \textbf{41.6} & \textbf{37.1}  \\ \hline
O & Oracle  & 38.7& 46.9& 56.7& 35.5& 49.4& 44.7& 35.9& 38.8& 43.1 \\ \hline
\end{tabular}}
\vskip -15.0pt
\end{table}

\subsection{Overall objective}
\vskip -5 pt
So far, we have introduced an Instance Relation Graph (IRG), Graph Distillation Loss (GDL), and Graph Contrastive Loss (GCL) to effectively tackle the source free domain adaptation problem for detection.
Then overall objective of our proposed SFDA method is formulated as:
\begin{equation}
\mathcal{L}_{SFDA} = \mathcal{L}_{SL}^{st} + \mathcal{L}_{GDL} + \mathcal{L}_{GCL},
\end{equation}


\section{Experiments and Results} \label{sec:experiments}
\vskip -2mm
To validate the effectiveness of our method, we compare our model performance with existing state-of-the-art UDA and SFDA methods on four different domain shift scenarios; 1) Adaptation to adverse weather, 2) Real to artistic, 3) Synthetic to real, and 4) Cross-camera.
Note that in UDA we have access to both source and target domain data. However, in SFDA, we have access only to source-trained model and not the source domain data for adaptation.

\vspace{-0.5 em}

\subsection{Implementation details}
Following the SFDA setting \cite{kim2021domain,li2020free}, we adopt Faster-RCNN \cite{ren2015faster} with ImageNet \cite{krizhevsky2012imagenet} pre-trained ResNet50 \cite{he2016deep} as the backbone. In all of our experiments, the input images are resized with a shorter side to be 600 while maintaining the aspect ratio and the batch size to 1.
The source model is trained using SGD optimizer with a learning rate of 0.001 and momentum of 0.9 for 10 epochs. For the proposed framework, the teacher network EMA momentum rate $\alpha$ is set equal to 0.9. In addition, the pseudo-labels generated by the teacher network with confidence greater than the threshold $T$=0.9 are selected for student training. We utilize the SGD optimizer to train the student network with a learning rate of 0.001 and momentum of 0.9 for 10 epochs. We report the mean Average Precision (mAP) with an IoU threshold of 0.5 for the teacher network on the target domain during the evaluation.

\vspace{-0.5 em}

\subsection{Quantitative comparison}

\subsubsection{Adaptation to adverse weather}
\noindent \textbf{Description.} Given a model trained on clear weather condition, we aim to perform adaptation to images in adverse weather conditions like fog/haze etc.
The Cityscapes \cite{cordts2016cityscapes} consist of 2,975 training images and  500 validation images with 8 object categories: \textit{person, rider, car, truck, bus, train, motorcycle and bicycle}.
The FoggyCityscapes \cite{Sakaridis2018SemanticFS} consist of images that are rendered from the Cityscapes dataset by integrating fog and depth information.
To this end, a model trained on Cityscapes is adapted to FoggyCityscapes without having access to the Cityscapes dataset.  

\noindent \textbf{Results.} Table~\ref{tab:foggy} provides the quantitative comparison with the existing UDA and SFDA methods for Cityscape$\rightarrow$FoggyCityscapes adaptation scenario.
From Table~\ref{tab:foggy}, we can infer that the proposed method outperforms most of the existing UDA methods such as SWDA \cite{saito2019strong}, InstanceDA \cite{wu2021instance}, and  CategoricalDA \cite{xu2020exploring}. However, compared MeGA-CDA \cite{vs2021mega} and Unbiased DA \cite{deng2021unbiased} methods, our proposed method produces a competitive performance with a drop of ~2.5-3.5 mAP. But it is worth noting that, these method make use of labelled source data during adaptation whereas our proposed method only has access to source-trained model.
Furthermore, compared with existing SFDA methods, SFOD \cite{li2020free} and HCL \cite{huang2021model}, the proposed method provides improvement of 4.3 mAP and 3.2 mAP, respectively.
We also compared with mean-teacher self-training baseline to show that adding the proposed GCL loss is able to enhance the features representation on the target domain, providing an improvement of 3.5 mAP.

\begin{table}[t]
\caption{ \small Quantitative results for  Sim10K $\rightarrow$  Cityscapes and  KITTI $\rightarrow$  Cityscapes.  S: Source only, UDA: Unsupervised Domain Adaptation, SFDA: Source-Free domain adaptation.}
\centering
\begin{center}
\label{tab:sim_kitti}
\vskip -15.0pt
\resizebox{0.98\linewidth}{!}{\begin{tabular}{clcc}
\hline
 Type  & Method & Sim10k $\rightarrow$ City & Kitti $\rightarrow$ City \\ \cline{3-4} 
                  &      & AP of Car      & AP of Car     \\ \hline
\hline
S  & Source Only   & 32.0          & 33.9  \\ \hline
 & DA Faster \cite{Chen2018DomainAF}    & 38.9          & 38.5   \\ 
 & Selective DA \cite{zhu2019adapting}   &  43.0& 42.5       \\
 & MAF \cite{he2019multi}   & 41.1 & 41.0\\
 & Robust DA \cite{khodabandeh2019robust}    & 42.5& 42.9\\
UDA & Strong-Weak \cite{saito2019strong} & 40.1& 37.9\\
  & ATF \cite{he2020domain}  &  42.8 & 42.1\\
  & Harmonizing \cite{chen2020harmonizing}  &  42.5 & -\\
  & Cycle DA \cite{zhao2020collaborative}  & 41.5 &  41.7\\  
  & MeGA CDA \cite{vs2021mega} & 44.8& 43.0\\
    & Unbiased DA \cite{deng2021unbiased} &43.1& -\\\hline
  &  SFOD \cite{li2020free} & 42.3& 43.6\\
  & SFOD-Mosaic \cite{li2020free}  & 42.9& 44.6\\
SFDA  & Mean-teacher \cite{tarvainen2017mean}    & 39.7 & 41.2 \\ 
  & IRG (Ours)    &  \textbf{45.2} & \textbf{46.9}\\ \hline
\end{tabular}}
\end{center}
\vskip -15.0pt
\end{table}

\begin{table}
\caption{\small Quantitative results for PASCAL-VOC $\rightarrow$  Watercolor.\hspace{\textwidth}S: Source only, UDA: Unsupervised Domain Adaptation, SFDA: Source-Free domain adaptation.}
\label{tab:water}
\huge
\centering
\begin{center}
\vskip -10.0pt
\resizebox{0.98\linewidth}{!}{\begin{tabular}{clccccccc}
\hline
Type  & Method         & bike          & bird        & car           & cat           & dog           & prsn          & mAP           \\ \hline

S  & Source only    & 68.8          & 46.8          & 37.2          & 32.7          & 21.3          & 60.7          & 44.6          \\\hline
  & DA Faster \cite{Chen2018DomainAF}     & 75.2 & 40.6          & 48.0          & 31.5 & 20.6          & 60.0          & 46.0          \\ 
    & BDC Faster \cite{saito2019strong}     & 68.6          & 48.3 & 47.2          & 26.5          & 21.7          & 60.5          & 45.5          \\ 
 & BSR \cite{kim2019self}     & 82.8 & 43.2          & 49.8          & 29.6 & 27.6          & 58.4          & 48.6          \\ 
UDA & WST \cite{kim2019self}   & 77.8 & 48.0          & 45.2          & 30.4 & 29.5          & 64.2          & 49.2          \\ 
  & SWDA \cite{saito2019strong}     & 71.3 & 52.0          & 46.6          & 36.2 & 29.2          & 67.3          & 50.4          \\ 
  & HTCN \cite{chen2020harmonizing}     & 78.6 & 47.5          & 45.6          & 35.4 & 31.0          & 62.2          & 50.1          \\ 

 & $\text{I}^{3}$Net \cite{chen2021i3net}    & 81.1 & 49.3          & 46.2          & 35.0 & 31.9          & 65.7          & 51.5          \\ 
 & Unbiased DA \cite{deng2021unbiased}    & 88.2 & 55.3          & 51.7          & 39.8 & 43.6          & 69.9          & 55.6          \\ \hline
    & PL \cite{khodabandeh2019robust}   & 74.6& 46.5& 45.1& 27.3& 25.9& 54.4& 46.1  \\ 
  & SFOD \cite{li2020free}    & \textbf{76.2}& 44.9& 49.3& \textbf{31.6}& 30.6& 55.2& 47.9  \\
SFDA  & Mean-teacher \cite{tarvainen2017mean}   & 73.6& 47.6& 46.6& 28.5& 29.4& 56.6& 47.1  \\ 
  & IRG (Ours)   & 75.9& \textbf{52.5}& \textbf{50.8}& 30.8& \textbf{38.7}& \textbf{69.2}& \textbf{53.0}  \\ \hline
\end{tabular}}
\end{center}

\vskip -20.0pt
\end{table}

\subsubsection{Realistic to artistic data adaptation}
\noindent \textbf{Description.} Here, we consider adaptation to dissimilar domains \cite{saito2019strong}, where a model trained on the real-world images is aimed to perform adaptation towards artistic domain.
We consider the model trained on the Pascal-VOC dataset \cite{everingham2010pascal} and adapt to two target domains, namely, Clipart \cite{inoue2018cross} and Watercolor \cite{inoue2018cross}.
The Clipart dataset contains 1K unlabeled images and has the same 20 categories as Pascal-VOC.
The Watercolor consists of 1K training and 1K testing images with six categories.

\noindent \textbf{Results.} The PASCAL-VOC$\rightarrow$Clipart adaptation results are reported in Table~\ref{tab:clip}.
Our method outperforms the existing UDA methods such as ADDA \cite{inoue2018cross} and DANN \cite{ganin2016domain} by a margin of 4.7 mAP and 0.3 mAP, respectively.
Moreover, the PASCAL-VOC$\rightarrow$Watercolor adaptation results are reported in Table~\ref{tab:water}.
Even in this case, our method outperforms the state-of-the-art UDA methods such as SWDA \cite{saito2019strong} and $\text{I}^{3}$Net \cite{chen2021i3net} by 2.6 mAP and 1.5 mAP, respectively.
Furthermore, for both Clipart and Watercolor adaptation scenarios, our method consistently outperforms in every category compared with pseudo-label self-training (PL) and mean-teacher baseline.

\begin{figure*}[t]
\begin{center}
\includegraphics[width=0.85\linewidth]{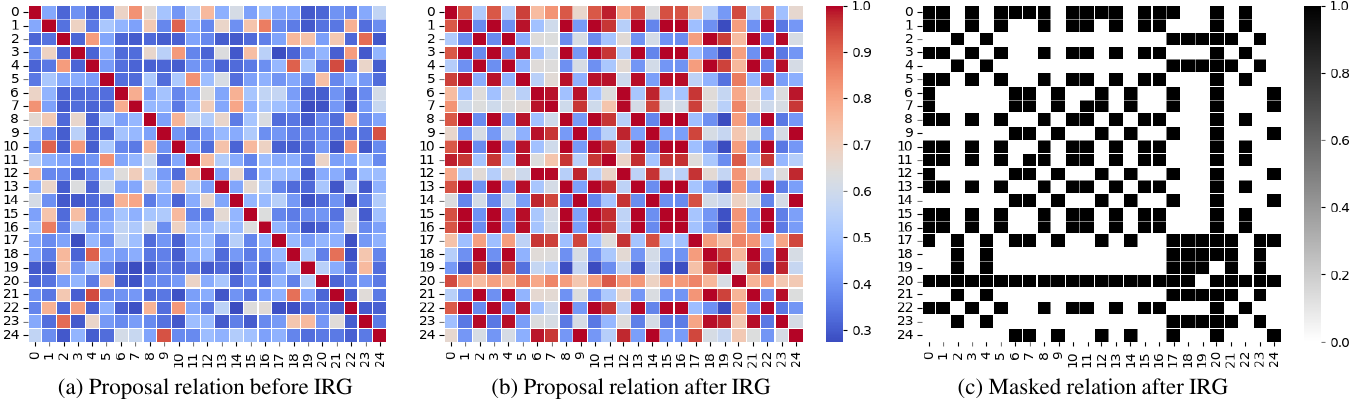}
\end{center}
 \vskip -15.0pt\caption{Relation matrix analysis for 25 proposal RoI features before and after passing through IRG network and corresponding masked instance pairwise labels. We can observe the IRG network models the relationship between the proposal, which maximizes the similarity between similar proposals and vice versa for dissimilar proposals. }
\label{fig:graph_vis}
\vskip -10.0pt
\end{figure*}

\begin{table*}[t]
\caption{ Quantitative results (mAP) for PASCAL-VOC $\rightarrow$  Clipart. S: Source only, UDA: Unsupervised Domain Adaptation, SFDA: Source-Free domain adaptation.}
\label{tab:clip}
\vskip -7.0pt
\huge
\centering
\resizebox{0.9\linewidth}{!}{\begin{tabular}{clccccccccccccccccccccc}
\hline
Type  & Method         & aero          & bcycle        & bird          & boat          & bottle        & bus           & car           & cat           & chair         & cow           & table         & dog           & horse         & bike          & prsn          & plnt          & sheep         & sofa          & train         & tv            & mAP  \\ \hline
\hline
S  & Source only    & 35.6          & 52.5          & 24.3          & 23.0          & 20.0          & 43.9          & 32.8          & 10.7          & 30.6          & 11.7          & 13.8          & 6.0           & 36.8          & 45.9          & 48.7          & 41.9          & 16.5          & 7.3           & 22.9          & 32.0          & 27.8 \\ \hline

    & DANN \cite{ganin2016domain}     & 24.1         & 52.6          & 27.5          & 18.5          & 20.3          & 59.3          & 37.4          & 3.8          & 35.1          & 32.6 & 23.9          & 13.8          & 22.5          & 50.9          & 49.9          & 36.3          & 11.6     & 31.3    & 48.0          & 35.8          & 31.8    \\

  & DAF \cite{Chen2018DomainAF}    & 15.0          & 34.6          & 12.4          & 11.9          & 19.8          & 21.1          & 23.3          & 3.10          & 22.1          & 26.3 & 10.6          & 10.0          & 19.6          & 39.4          & 34.6          & 29.3          & 1.00          & 17.1          & 19.7          & 24.8          & 19.8 \\ 
    & ADDA \cite{inoue2018cross}    & 20.1          & 50.2          & 20.5          & 23.6          & 11.4          & 40.5          & 34.9          & 2.3          & 39.7          & 22.3 & 27.1          & 10.4          & 31.7          & 53.6          & 46.6          & 32.1          & 18.0         & 21.1          & 23.6          & 18.3          & 27.4 \\ 
  UDA   & BDC Faster \cite{saito2019strong}   & 20.2          & 46.4          & 20.4          & 19.3          & 18.7          & 41.3          & 26.5          & 6.40          & 33.2          & 11.7          & 26.0 & 1.7           & 36.6 & 41.5          & 37.7          & 44.5         & 10.6          & 20.4          & 33.3          & 15.5          & 25.6 \\ 
     & CRDA \cite{xu2020exploring}  & 28.7 & 55.3 & 31.8 & 26.0 & 40.1 & 63.6 & 36.6 & 9.4 & 38.7 & 49.3 & 17.6 & 14.1 & 33.3 & 74.3 & 61.3 & 46.3 & 22.3 & 24.3 & 49.1 & 44.3 & 38.3 \\ 
     & Unbiased DA \cite{deng2021unbiased}   & 39.5&60.0&30.5&39.7&37.5&56.0&42.7&11.1&49.6&59.5&21.0&29.2&49.5&71.9&66.4&48.0&21.2&13.5&38.8&50.4&41.8\\ \hline
     & PL \cite{tarvainen2017mean}  & 18.3          & 48.4          & 19.2          & 22.4          & 12.8          & 38.9          & 36.1          & 5.2          & 36.9          & 24.8 & \textbf{29.3}          & 9.09          & 34.6          & 58.6          & 43.1          & 34.3          & 9.09         & 14.4          & 26.9          & 19.8          & 28.2 \\ 
  & SFOD  \cite{li2020free}  & 20.1& 51.5& 26.8& \textbf{23.0}& 24.8& 64.1& \textbf{37.6}& 10.3&  36.3& 20.0& 18.7& \textbf{13.5}& 26.5& 49.1& 37.1& 32.1& \textbf{10.1}& 17.6& 42.6& 30.0& 29.5 \\
SFDA  & Mean-teacher \cite{tarvainen2017mean}   & \textbf{22.3}& 42.3& 23.8& 21.7& 23.5& 60.7& 33.2& 9.1& 24.7 & \textbf{16.7}& 12.2& 13.1& 26.8& \textbf{73.6}& 43.9& 34.5& 9.09& 24.3 & 37.9& 42.2& 29.1  \\ 
  & IRG (Ours)   & 20.3& \textbf{47.3}& \textbf{27.3}& 19.7& \textbf{30.5}& \textbf{54.2}& 36.2& \textbf{10.3}&  \textbf{35.1}& 20.6& 20.2& 12.3& \textbf{28.7}& 53.1& \textbf{47.5}& \textbf{42.4}& 9.09& 21.1& \textbf{42.3}& \textbf{50.3}& \textbf{31.5} \\ \hline
\end{tabular}}
\vskip -15.0pt
\end{table*}

\subsubsection{Synthetic to real-world adaptation}
\noindent \textbf{Description.} The cost of generating and labeling synthetic data is low compared to real-world data.
Hence, it makes sense to train a detector on synthetic images and transfer the knowledge to real-world data.
However, the style shift between synthetic to real domain makes it challenging.
Here, we consider such scenario where we adapt a model trained on the synthetic data, Sim10K \cite{johnson2016driving}, to a real-world data, Cityscapes \cite{cordts2016cityscapes} under SFDA condition, i.e., synthetic data are not available while adapting the model to the real-world images.
The model is trained on 10,000 training images of Sim10k rendered by the \textit{Grand Theft Auto} gaming engine.
The target Cityscapes dataset consists of 2,975 unlabeled training images and 500 validation images. 

\noindent \textbf{Results.} We report the results of Sim10K$\rightarrow$Cityscapes in Table~\ref{tab:sim_kitti}.
Note that even though we adapt for only the car category, the proposed GCL training strategy is able to get discriminative positive pairs for different cars and improve the feature representations through contrastive training. Our proposed method outperforms existing UDA method like Cycle DA \cite{zhao2020collaborative}, Unbiased DA \cite{deng2021unbiased} etc. by considerable margin. Under SFDA setting, the proposed method produces state-of-the-art performance by improving $\sim$1 mAP compared to  SFOD \cite{li2020free}.

\subsubsection{Cross-camera adaptation} 
\noindent \textbf{Description.}
In real-world scenarios, the target domain data is captured by a camera with configurations different from the source data.
To emulate this cross-camera conditions, we consider a model trained on source, KITTI dataset \cite{geiger2013vision}, is adapted to target, Cityscapes \cite{cordts2016cityscapes}.
The KITTI dataset consists of 7,481 training images, which is used to get the source-trained detector model.
The model is then adapted to the target domain dataset, i.e., Cityscapes.

\noindent \textbf{Results.} KITTI$\rightarrow$Cityscapes results are reported in Table~\ref{tab:sim_kitti}.
Our method outperform existing state-of-the-art UDA methods like Cycle DA \cite{zhao2020collaborative}, MeGA CDA \cite{vs2021mega} and Unbiased DA \cite{deng2021unbiased} by considerable margin. Further in SFDA setting, the proposed method produce state-of-the-art performance by improving around 1.1 mAP compared to SFOD.

\subsection{Ablation analysis}
We study the impact of the proposed GCL and IRG network by performing an in-depth ablation analysis on Cityscapes$\rightarrow$FoggyCityscapes adaptation scenario. 

\noindent\textbf{Quantitative analysis.}  The results for Cityscapes$\rightarrow$FoggyCityscapes ablation experiments are reported in Table~\ref{foggy_ab}.
In Table~\ref{foggy_ab}, the first three experiments are performed to analyze the effect of various combinations of weak and strong augmentation for a mean-teacher framework in an SFDA setting.
More precisely, we input the student and teacher network with \textit{Weak-Weak (WW)}, \textit{Strong-Strong (SS)} and \textit{Strong-Weak (SW)} augmented images, respectively.
These three experiments show that \textit{strong-weak (SW)} produces consistent and improved results compared to other variations.
This is due to mutual learning between student and teacher networks, where student trains on strong augmentation leading to robust prediction and the teacher supervise the student by good pseudo-labels predicted from the weak augmented images.
Furthermore, minimizing the \textit{discrepancy between instance relation graph} network of student and teacher framework ensures consistency between student and teacher graph proposal feature representations.
Subsequently, addition of \textit{graph distillation loss} enhances the model performance from 34.3 mAP to 35.9 mAP.
Finally, utilizing \textit{graph-guided contrastive learning} on the proposal features further helps the model learn high-quality representations, resulting in an increase in performance by 1.9 mAP on the target domain.

\noindent\textbf{Qualitative analysis.} In Fig.~\ref{fig:graph_vis}, we show the relation matrix for the RoI features before and after it is processed by IRG.
For better visualizations, we consider 25 out of 300 RoI features.
It can be observed that relation between the proposals are poorly defined and IRG network is able to improve these relations through graph-based feature aggregation.

\begin{table}[!t]
\caption{Ablation study on FoggyCityscapes.}
\vskip -10.0pt
\centering
\huge
\resizebox{0.95\linewidth}{!}{
\begin{tabular}{lcccccccccccc} \hline
Method & PL & GDL & GCL & prsn & rider & car & truc & bus & train & mcycle & bcycle & mAP \\ \hline
Source Only & \xmark & \xmark & \xmark & 25.8& 33.7& 35.2& 13.0& 28.2& 9.1& 18.7& 31.4& 24.4\\
\hline
MT + WW & \cmark & \xmark & \xmark  & 35.8  & 42.6 & 43.9 & 23.1 & 32.7 & 11.0 & 29.9 & 38.7 & 32.2  \\ 
MT + SS & \cmark & \xmark & \xmark  & 32.8  & 41.4 & 43.8 & 18.2 & 28.6 & 11.2 & 24.6 & 38.3 & 29.9  \\ 
MT + SW & \cmark & \xmark & \xmark  & 33.9  & 43.0 & 45.0 & \textbf{29.1} & 37.2 & 25.1 & 25.5 & 38.2 & 34.3  \\ 
Ours    & \cmark & \cmark & \xmark  & 37.2  & 43.1 & 51.0 & 28.6 & \textbf{40.1} & 21.2 & 28.2 & 37.1 & 35.9  \\ 
Ours    & \cmark & \cmark & \cmark  & \textbf{37.4}& \textbf{45.2}& \textbf{51.9}& 24.4& 39.6& \textbf{25.2}& \textbf{31.5}&  \textbf{41.6} & \textbf{37.1}  \\ \hline
\end{tabular}}
\label{foggy_ab}
\vskip -15.0pt
\end{table}

\vspace{-0.5 em}
\section{Conclusion}

In this work, we presented a novel approach for source-free domain adaptive detection using graph-guided contrastive learning.
Specifically, we introduced a contrastive graph loss to enhance the target domain representations by exploiting instance relations.
We propose an instance relation graph network built on top of a graph convolution network to model the relation between proposal instances.
Subsequently, the learned instance relations are used to get positive/negative proposal pairs to guide contrastive learning.
We conduct extensive experiments on multiple detection benchmarks to show that the proposed method efficiently adapts a source-trained object detector to the target domain, outperforming the state-of-the-art source-free domain adaptation and many unsupervised domain adaptation methods.
%
%
{\small
\bibliographystyle{splncs04}
\bibliography{egbib}
}

\newpage

\onecolumn

\begin{center}
 {\huge Supplementary material: Instance Relation Graph \\ Guided Source-Free Domain Adaptive Object Detection}
\end{center}

\noindent \textbf{Ablation analysis for different loss functions:} In Table~\ref{foggy_ab_con}, we present more ablation analysis for Cityscapes$\rightarrow$FoggyCityscapes experiment. In Table~\ref{foggy_ab_con}, the first experiment is standard mean-teacher \cite{tarvainen2017mean} framework for SFDA setting. The second experiment is performed to understand the effect between supervised and our proposed contrastive loss. The supervised contrastive loss \cite{khosla2020supervised} is computed between the proposals generated from the student and teacher network. The class information for each proposal is obtained after passing through the RoI feature extractor and classification head. Using the proposal features and class information between student and teacher networks, we compute the SupCon loss. Therefore compared to mean-teacher training, supervised contrastive learning on mean-teacher increase the performance by 1.9 mAP. Without supervised contrastive learning, distillation loss between student and teacher networks degrades the performance from 36.2 mAP to 35.9 mAP. In final, utilizing graph-guided contrastive learning on the proposal features on top of GDL further helps the model learn high-quality representations, increasing performance by 1.9 mAP on the target domain. Note all augmentations are applied on the image level. Also, strong and weak augmentation essentially simulates the domain gap between student and teacher. Exploiting this property and mean-teacher framework, the consistency loss between strongly augmented student prediction and weakly augmented teacher predictions enforces the student network to learn a more robust and domain-invariant feature representation. Strong augmentation: color jitter, grayscale, Gaussian blur, erasing. Weak augmentation: horizontal flip.

\begin{table}[H]
\caption{Ablation study for different loss functions.}
\vskip -5.0pt
\centering
\huge
\resizebox{0.7\linewidth}{!}{
\begin{tabular}[!t]{lccccccccccccc} \toprule
Method & PL & SimCLR & GDL & GCL & prsn & rider & car & truc & bus & train & mcycle & bcycle & mAP \\ \toprule
Source Only & \xmark & \xmark & \xmark & \xmark & 25.8& 33.7& 35.2& 13.0& 28.2& 9.1& 18.7& 31.4& 24.4\\
\toprule
MT + SW & \cmark & \xmark & \xmark & \xmark & 33.9  & 43.0 & 45.0 & \textbf{29.1} & 37.2 & 25.1 & 25.5 & 38.2 & 34.3  \\ 
MT + SW & \cmark & \cmark & \xmark & \xmark & 36.1  & \textbf{45.8} & 47.2 & 28.5 & 36.6 & 29.5 & 27.2 & 38.8 & 36.2  \\ 
Ours    & \cmark & \xmark & \cmark & \xmark & 37.2  & 43.1 & 51.0 & 28.6 & 40.1 & 21.2 & 28.2 & 37.1 & 35.9  \\ 
Ours    & \cmark & \xmark & \cmark & \cmark & \textbf{37.4}. & 45.2 & \textbf{51.9} & 24.4 & \textbf{39.6} & \textbf{25.2} & \textbf{31.5} & \textbf{41.6} & \textbf{37.1}  \\ \toprule
\end{tabular}
}
\vskip -15.0pt
\label{foggy_ab_con}
\end{table}

\noindent \textbf{Ablations analysis on IRG network}: In Table \ref{tab:all_ab}, in the second row, we use Kmeans clustering algorithm to find positive proposal pairs instead of IRG network. Specifically, we utilize kmeans algorithm to find positive proposals and then apply CRL loss. From Table \ref{tab:all_ab}, we can infer that using learnable IRG to model positive relations is more effective than kmeans and improves the performance by considerable margin. In a next row, we freeze the IRG network and use the original edge weights to compute CRL loss. In frozen IRG experiment, when we use original edge weights to compute GCL loss the performance drops to 36.2 mAP compared to learnable IRG network.

\vskip -12.5pt
\begin{table}[H]
\caption{Ablation study on IRG network}
\vskip -7.5pt
\centering
\huge
\resizebox{0.7\linewidth}{!}{
\begin{tabular}{lcccccccccccc} \hline
Method & CRL & prsn & rider & car & truc & bus & train & mcycle & bcycle & mAP \\ \hline
Source Only & \xmark & 25.8& 33.7& 35.2& 13.0& 28.2& 9.1& 18.7& 31.4& 24.4\\
Kmeans  & \cmark & 36.3 & 44.5 & 49.7 & 26.2 & 37.9 & 26.0 & 32.8 & 39.3  & 36.5  \\ 
IRG(Frozen)  & \cmark & 33.9 & 43.7 & 47.3 & 26.8 & 38.5 & 27.1 & 30.2 & 38.9  & 36.2  \\ 
Ours & \cmark & 37.4& 45.2& 51.9 & 24.4& 39.6 & 25.2 & 31.5 & 41.6 & 37.1  \\ \hline
\end{tabular}}
\label{tab:all_ab}
\vskip -15.0pt
\end{table}

\begin{figure}[H]
\begin{center}
\includegraphics[width=0.6\linewidth]{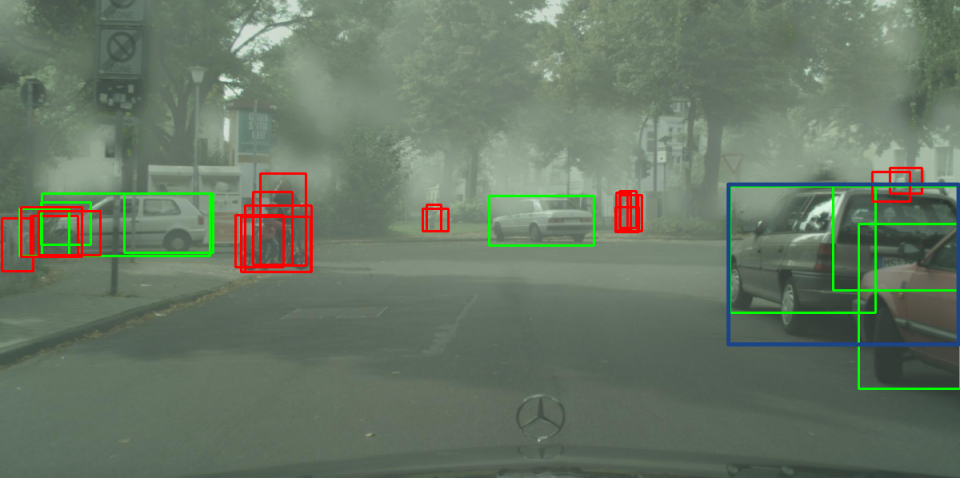}
\end{center}
\caption{Blue: Query proposal, Green: positive proposals, Red: Negative proposals}
\label{fig:positive}
\end{figure}

\noindent \textbf{Positive proposals visualization}: From Fig \ref{fig:positive}, given a anchor (blue), we visualized it's positive (green) and negative (red) proposals generated by the IRG network.

\begin{figure*}
\begin{center}
\includegraphics[width=0.45\linewidth]{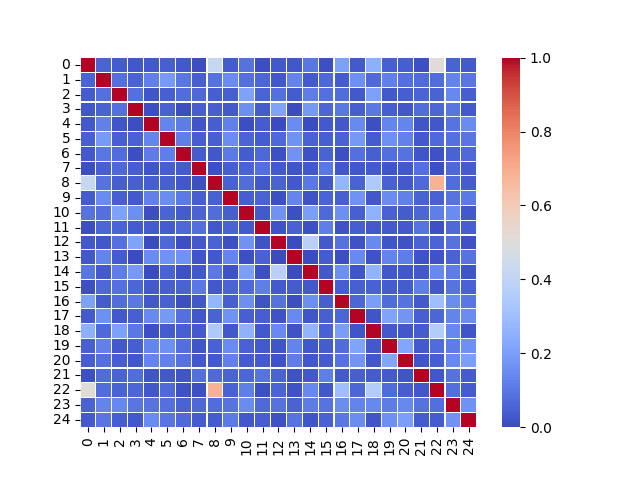}
\includegraphics[width=0.45\linewidth]{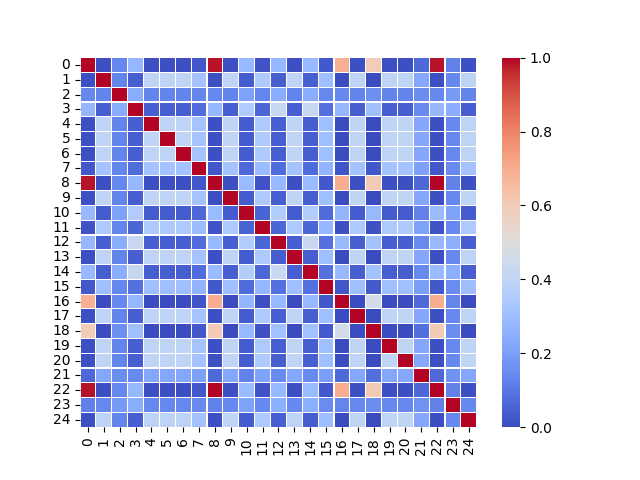}
\includegraphics[width=0.45\linewidth]{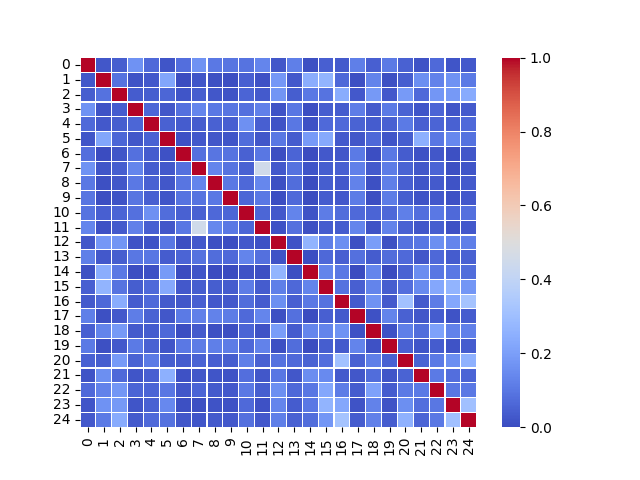}
\includegraphics[width=0.45\linewidth]{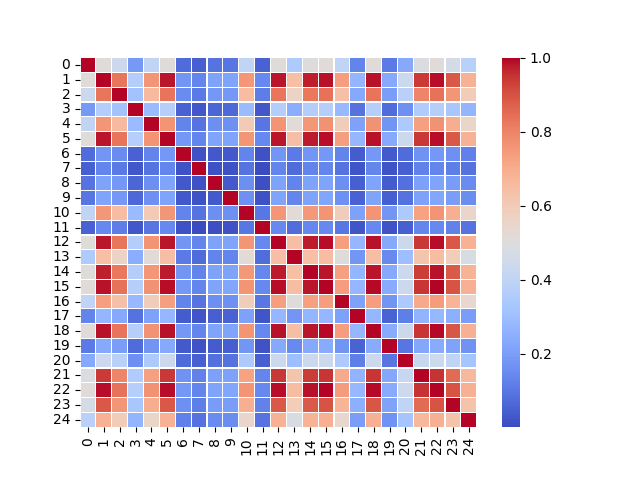}
\includegraphics[width=0.45\linewidth]{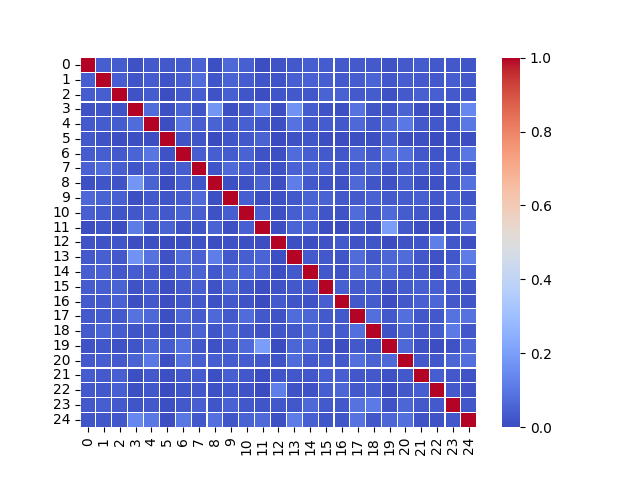}
\includegraphics[width=0.45\linewidth]{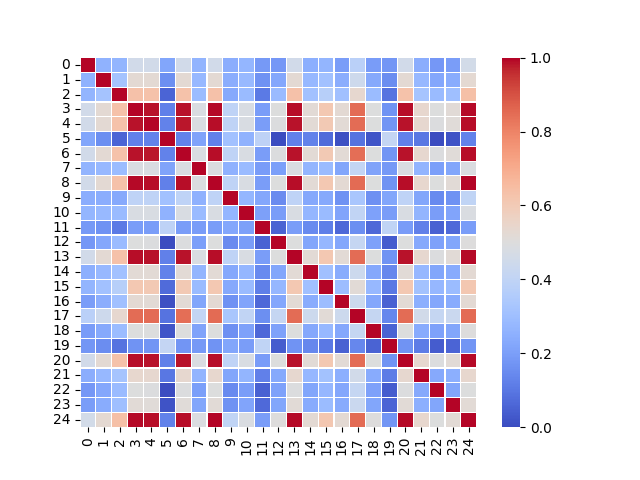}
\includegraphics[width=0.45\linewidth]{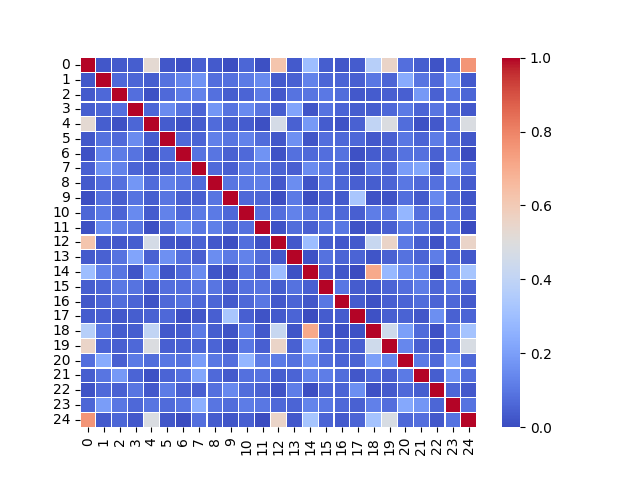}
\includegraphics[width=0.45\linewidth]{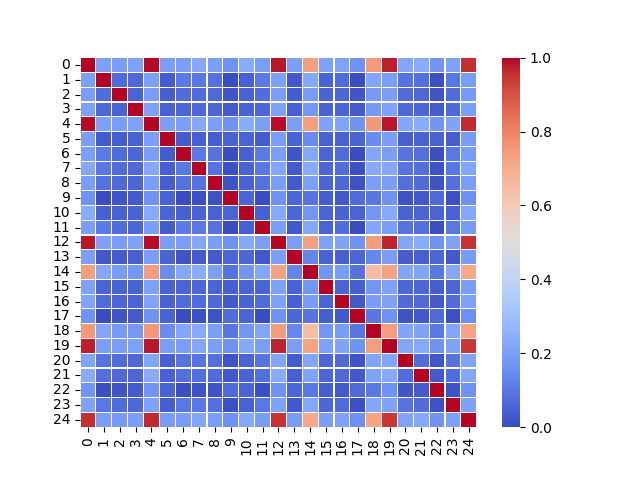}

\end{center}
\vskip -10.0pt \caption{More visualizations of relation matrix of RoI features before and after passing through IRG network.}
\label{fig:sup_graph_vis}
\end{figure*}

\begin{figure*}
\begin{center}
\includegraphics[width=0.45\linewidth]{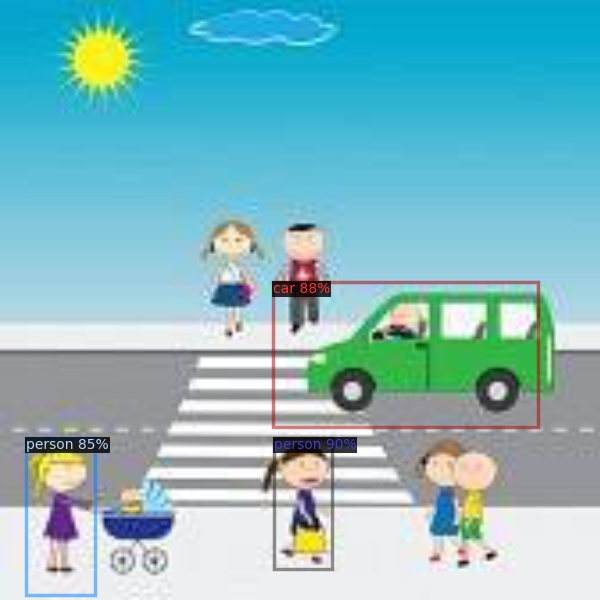}
\includegraphics[width=0.45\linewidth]{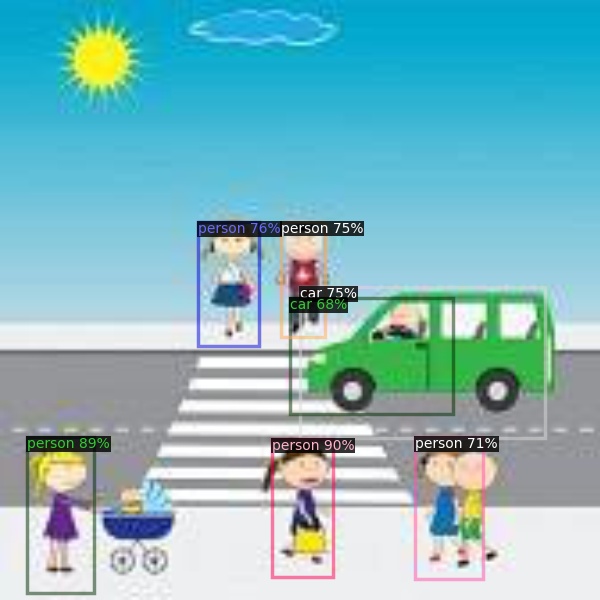}
\includegraphics[width=0.45\linewidth]{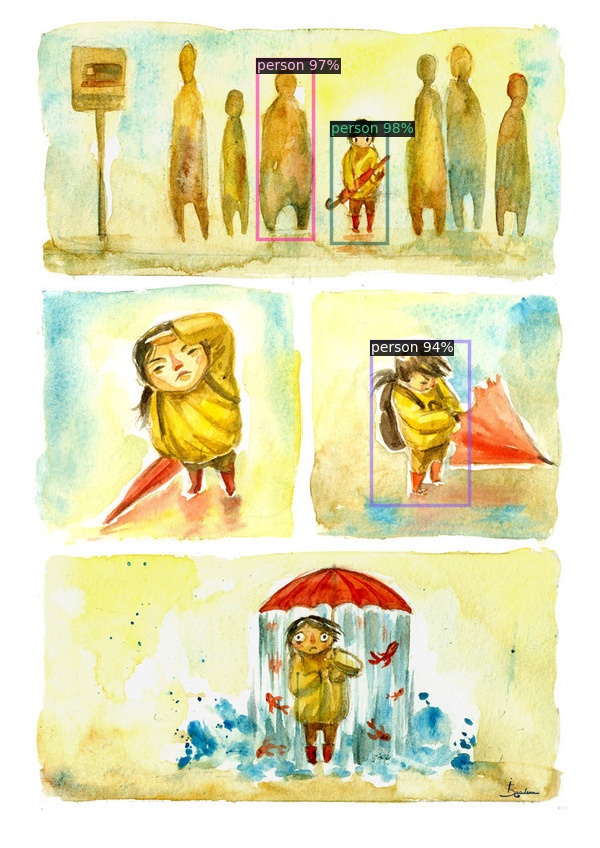}
\includegraphics[width=0.45\linewidth]{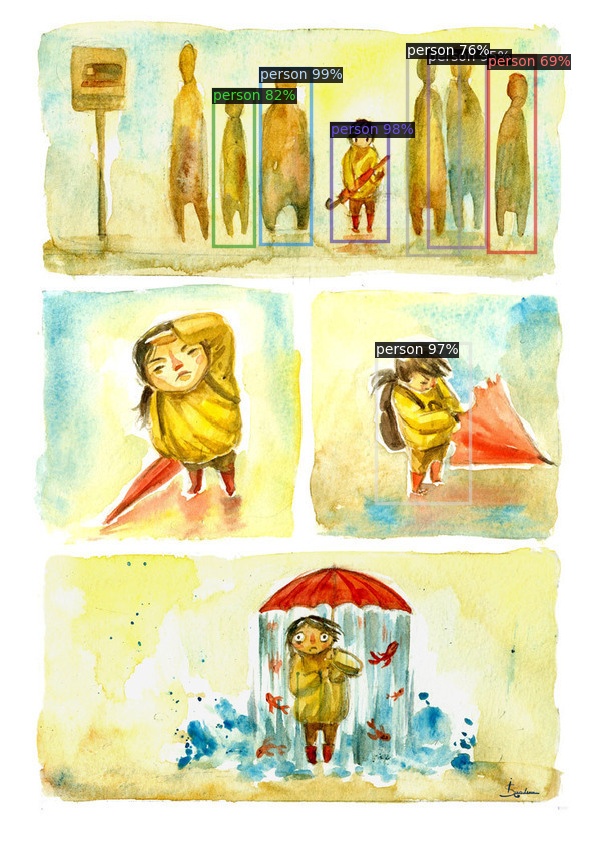}
\vskip -5pt
 Student-teacher \hskip 200pt Ours
 
\end{center}
\vskip -10.0pt \caption{More detection visualization for realistic to artistic adaptation.  From the above visualization, we can infer that our model efficiently tackles classes' negative transfer and constructs high confidence prediction boxes.}
\label{fig:sup_graph_vis}
\end{figure*}

\begin{figure*}
\begin{center}

\includegraphics[width=0.45\linewidth]{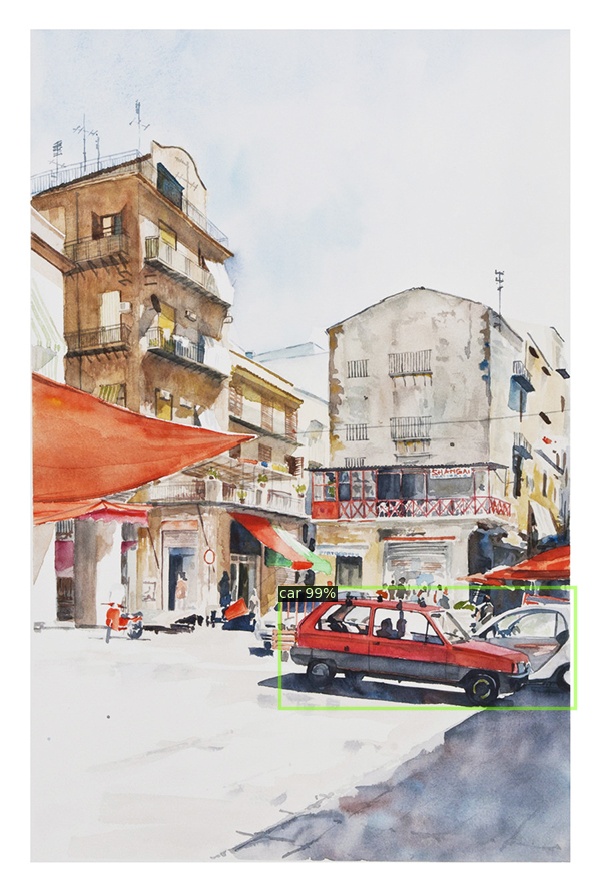}
\includegraphics[width=0.45\linewidth]{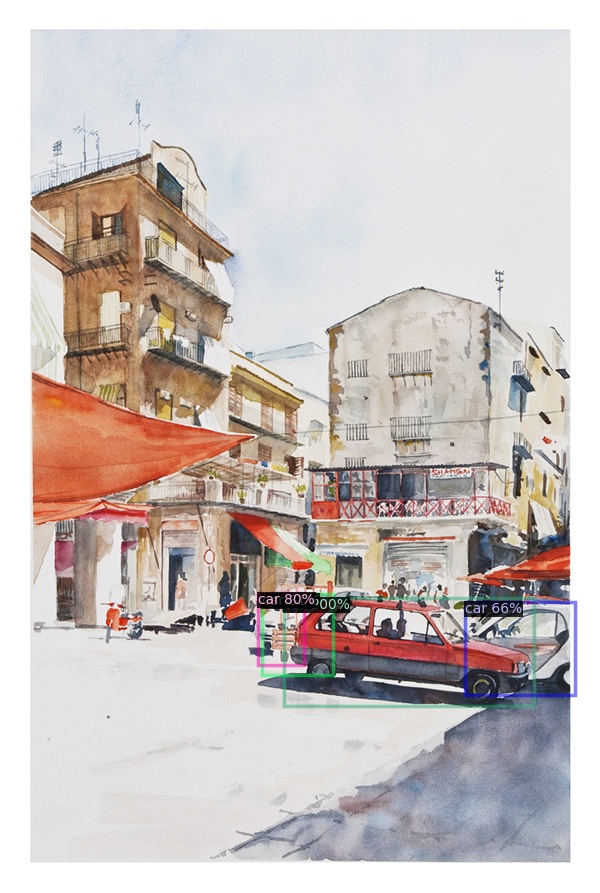}
\includegraphics[width=0.45\linewidth]{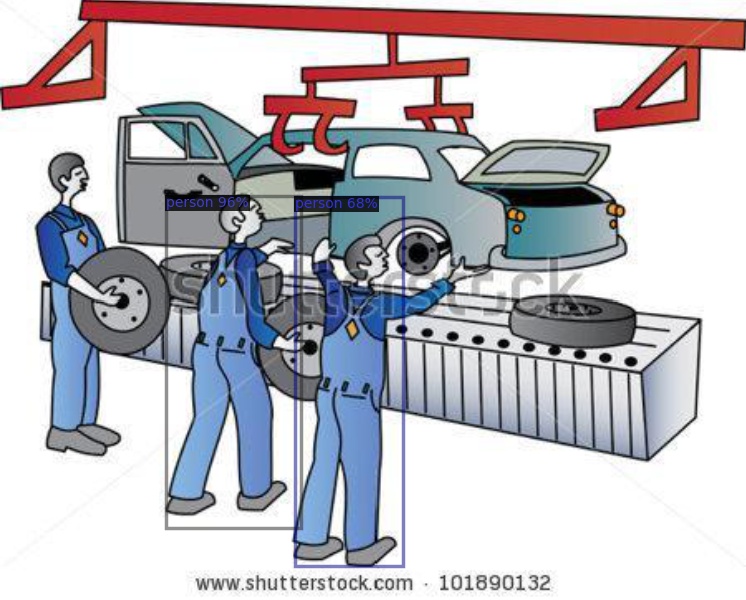}
\includegraphics[width=0.45\linewidth]{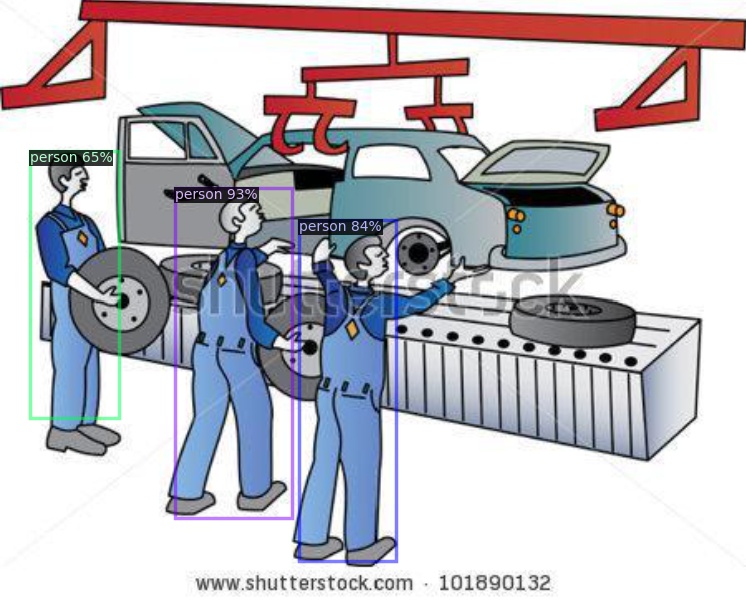}
\vskip -5pt
 Student-teacher \hskip 200pt Ours 
 
\end{center}
\vskip -10.0pt \caption{More detection visualization for realistic to artistic adaptation.  From the above visualization, we can infer that RPN fails to generate proposals due to tiny/occluded objects or heavy domain shifts, those instances cannot take full benefits of the proposed contrastive learning strategy.}
\label{fig:sup_graph_vis}
\end{figure*}

\end{document}